\documentclass[times,doublespace]{oupau}%For paper submission
\usepackage[utf8]{inputenc}
\usepackage{graphicx}
\usepackage{subcaption}
\usepackage{hyperref}

\makeatletter
\let\@received\@empty
\makeatother

\begin{document}

\runningheads{L. Melo}{Evaluation of Attention Mechanisms in U-Net}

\title{Evaluation of Attention Mechanisms in U-Net Architectures for Semantic Segmentation of Brazilian Rock Art Petroglyphs}

\author{Leonardi Melo\affil{a,b}\corrauth,
Luis Gustavo\affil{a}, Dimmy Magalhães\affil{b}, Lucciani Vieira\affil{b}, and Mauro Araújo\affil{b}}

\address{\affilnum{a}Computer Vision Research Center, iCEV Institute of Higher Education, Teresina, Brazil.\\
\affilnum{b}iCEV Institute of Higher Education, Teresina, Brazil.}
\corraddr{joao\_leonardi.melo@somosicev.com}

\begin{abstract}
This study presents a comparative analysis of three U-Net-based architectures for semantic segmentation of rock art petroglyphs from Brazilian archaeological sites. The investigated architectures were: (1) BEGL-UNet with Border-Enhanced Gaussian Loss function; (2) Attention-Residual BEGL-UNet, incorporating residual blocks and gated attention mechanisms; and (3) Spatial Channel Attention BEGL-UNet, which employs spatial-channel attention modules based on Convolutional Block Attention Module. All implementations employed the BEGL loss function combining binary cross-entropy with Gaussian edge enhancement. Experiments were conducted on images from the Poço da Bebidinha Archaeological Complex, Piauí, Brazil, using 5-fold cross-validation. Among the architectures, Attention-Residual BEGL-UNet achieved the best overall performance with Dice Score of 0.710, validation loss of 0.067, and highest recall of 0.854. Spatial Channel Attention BEGL-UNet obtained comparable performance with DSC of 0.707 and recall of 0.857. The baseline BEGL-UNet registered DSC of 0.690. These results demonstrate the effectiveness of attention mechanisms for archaeological heritage digital preservation, with Dice Score improvements of 2.5-2.9\% over the baseline.
\end{abstract}

\keywords{Semantic segmentation, Rock art, Attention mechanisms, Archaeological digital heritage, Deep learning}

% Submission dates - uncomment and fill in when submitting to journal
%\received{DD Month YYYY}
%\revised{DD Month YYYY}
%\accepted{DD Month YYYY}

\maketitle

\section{Introduction}
Digital preservation of archaeological sites has become increasingly relevant due to natural and anthropogenic degradation affecting rock art heritage. In northeastern Brazil, thousands of sites are registered by the Brazilian National Historic and Artistic Heritage Institute (IPHAN) \cite{IPHAN2023}, including Serra da Capivara National Park (Piauí) with 1,354 documented sites, with over one thousand sites containing rock art manifestations \cite{FUMDHAM2023,SerraCapivara2023}. The Poço da Bebidinha Archaeological Complex, located in Buriti dos Montes (Piauí), contains low-relief petroglyphs engraved on paragneiss outcrops of the Canindé Unit. These petroglyphs are distributed across approximately 2 km along the Poti River margins and exhibit geometric, zoomorphic, and anthropomorphic motifs associated with the Nordeste and Agreste traditions \cite{Viana2013Thesis,Conservacao2020,Ferreira2018Thesis}.

Semantic segmentation of rock art presents technical challenges arising from: (1) low contrast between engravings and rock substrate; (2) differential weathering degradation; (3) presence of lichens and mineral deposits; (4) illumination variations; and (5) temporal superposition of engravings \cite{horn2021}. Traditional methods based on adaptive thresholding and edge detection exhibit limitations, with false positive rates exceeding 40\% \cite{seidl2012}.

Convolutional neural networks such as the U-Net architecture \cite{ronneberger2015} represent state-of-the-art approaches in semantic segmentation. Bai et al. \cite{bai2023} introduced the Border-Enhanced Gaussian Loss (BEGL) function for rock art segmentation, combining binary cross-entropy with Gaussian edge enhancement. The framework achieved 0.865 Dice Similarity Coefficient (DSC) on the 3D-pitoti dataset \cite{poier2017pitoti}. Comparative studies of architectural variations incorporating attention mechanisms on Brazilian archaeological datasets remain limited.

Archaeological sites in Brazil exhibit geological, geomorphological, and stylistic characteristics that differ from European comparative collections such as the 3D-pitoti dataset \cite{poier2017pitoti}. The Poço da Bebidinha Complex presents mineralogical composition and erosion patterns distinct from those observed in European limestone caves. The Nordeste tradition of Brazilian rock art incorporates geometric motifs and superposition forms that differ from representations and stylistic conventions associated with European Paleolithic art.

The contributions of this work are threefold. First, it investigated deep neural networks with attention mechanisms for rock art segmentation on petroglyphs from the Poço da Bebidinha Complex. Second, it developed an annotated dataset of Brazilian petroglyphs for digital archaeological heritage documentation. Third, it provided experimental evaluation comparing different attention mechanism architectures (gated attention with residual blocks vs. CBAM-based spatial-channel attention) for rock art segmentation.

The remainder of this paper is organized as follows: Section 2 reviews related work on rock art segmentation and U-Net architectures; Section 3 describes the experimental methodology, including dataset characteristics, implemented architectures, and training protocol; Section 4 reports quantitative results and qualitative analysis; Section 5 discusses findings and practical implications; and Section 6 presents conclusions and future directions.

\section{Related Work}

\subsection{Rock Art Segmentation}

Computational methods for rock art analysis were introduced in the 1990s with image processing techniques based on morphological filters \cite{bednarik1994}. Subsequent methods included Hough transform for line detection \cite{desolneux2008}, principal component analysis for dimensionality reduction \cite{fritz2016}, and k-means clustering for pattern grouping \cite{mourao2016}.

Zeppelzauer et al. \cite{zeppelzauer2016} applied Support Vector Machines with Haar-like features for petroglyph classification, achieving 78.4\% accuracy on a dataset of 1,847 images. More recent applications employ deep learning with Convolutional Neural Networks (CNNs) for classification. Jalandoni et al. \cite{jalandoni2022} utilized machine learning models for automatic identification of painted rock art, achieving 89\% accuracy. Defrasne et al. \cite{defrasne2021} applied digital documentation techniques for analysis of Paleolithic rock art.

\subsection{U-Net Architecture and Variants}

The U-Net architecture, proposed by Ronneberger et al. \cite{ronneberger2015}, established a reference framework for semantic segmentation through its symmetric encoder-decoder architecture with skip connections. The structure enables the preservation of high-resolution spatial information during the segmentation process, utilizing 3×3 convolutions followed by batch normalization and ReLU activation \cite{ioffe2015batch}.

He et al. \cite{he2016} introduced residual blocks that facilitated training of deeper networks through identity connections. These blocks were subsequently refined with pre-activation design \cite{he2016identity} for improved gradient flow. Integration of residual blocks in U-Net architectures enables more stable training and improved convergence characteristics.

Oktay et al. \cite{oktay2018} introduced the Attention U-Net, incorporating attention gates for automatic suppression of irrelevant features.

\subsection{Attention Mechanisms}
Attention mechanisms were proposed by Bahdanau et al. \cite{bahdanau2014neural} for neural machine translation and extended by Vaswani et al. \cite{vaswani2017attention} with self-attention. In computer vision, attention enables differential weighting of spatial or channel features.

Hu et al. \cite{hu2018squeeze} proposed Squeeze-and-Excitation Networks (SE-Net) with channel attention via global average pooling and two fully connected layers. Wang et al. \cite{wang2020eca} presented ECA-Net (Efficient Channel Attention), without dimensional reduction, utilizing 1D convolution with adaptive kernel.

Woo et al. \cite{woo2018} proposed CBAM (Convolutional Block Attention Module), combining channel and spatial attention sequentially. The channel attention module utilizes dual pooling (average and max) for spatial aggregation:

\begin{align}
\mathbf{M}_c(\mathbf{F}) &= \sigma(\text{MLP}(\text{AvgPool}(\mathbf{F})) + \text{MLP}(\text{MaxPool}(\mathbf{F}))) \label{eq:cbam_channel}
\end{align}

The spatial module processes channel statistics with 7×7 convolution:

\begin{align}
\mathbf{M}_s(\mathbf{F}) &= \sigma(\text{Conv}^{7 \times 7}([\text{AvgPool}(\mathbf{F}); \text{MaxPool}(\mathbf{F})])) \label{eq:cbam_spatial}
\end{align}
\section{MATERIALS AND METHODS}

\subsection{Experimental Design}
This comparative study was designed to evaluate three U-Net-based architectures with BEGL loss function on a Brazilian petroglyph dataset. The experimental design comprised: (1) collection and manual annotation of 82 images from the Poço da Bebidinha site; (2) 6× data augmentation applied to the training images; (3) training of three architectures (BEGL-UNet baseline, Attention-Residual BEGL-UNet, Spatial-Channel Attention BEGL-UNet) with identical hyperparameters using 5-fold cross-validation; and (4) quantitative and qualitative evaluation.

\subsection{Statistical Analysis}
Evaluated metrics included Dice Score, Validation Loss, Precision, Recall, F1-Score, and Pixel Accuracy. Mean values across 5-fold cross-validation were computed for all architectures. Due to the limited sample size ($N = 5$ folds), statistical significance testing was not performed, and results are presented as descriptive comparisons of model performance.

\subsection{Dataset}

The dataset comprises 82 images from the Poço da Bebidinha archaeological site (Piauí) with segmentation masks. Images were resized and cropped to $512 \times 512$ pixels according to neural network input requirements. Preprocessing applied min-max intensity normalization to the [0,1] range.

\begin{figure}[h]
\centering
\begin{subfigure}{0.18\textwidth}
\centering
\includegraphics[width=\textwidth]{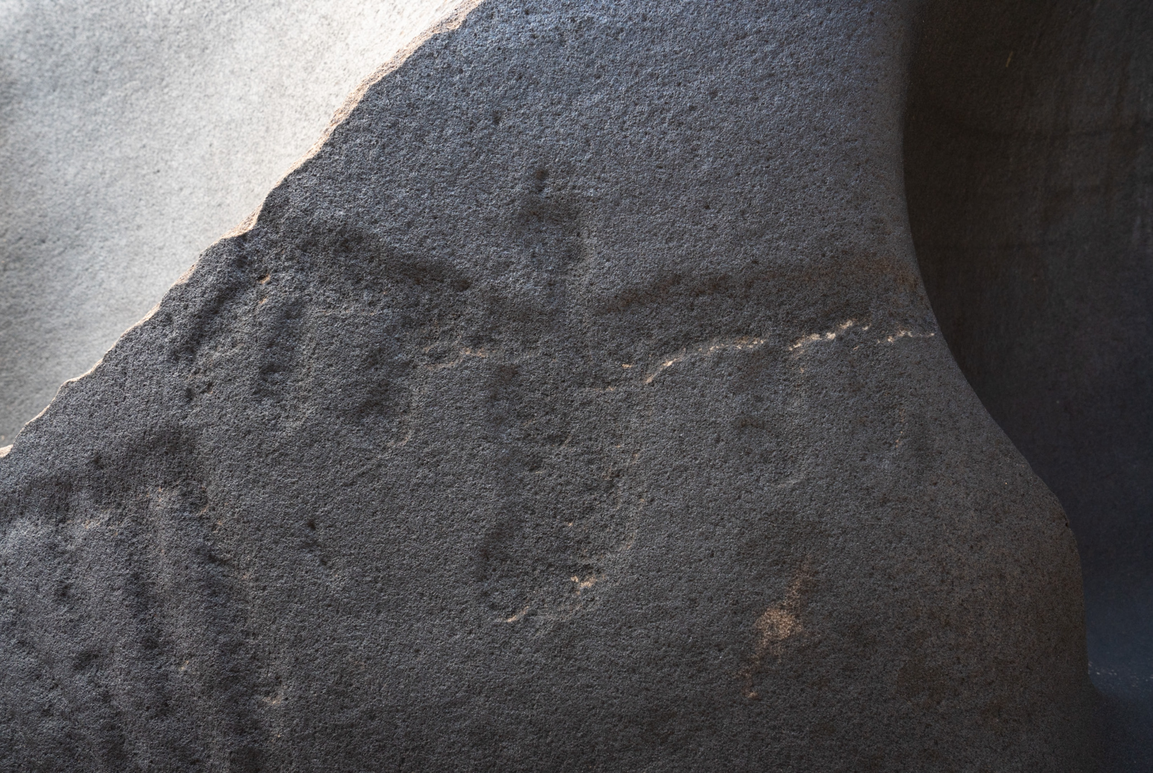}
\caption{}
\label{fig:sample10}
\end{subfigure}
\hfill
\begin{subfigure}{0.18\textwidth}
\centering
\includegraphics[width=\textwidth]{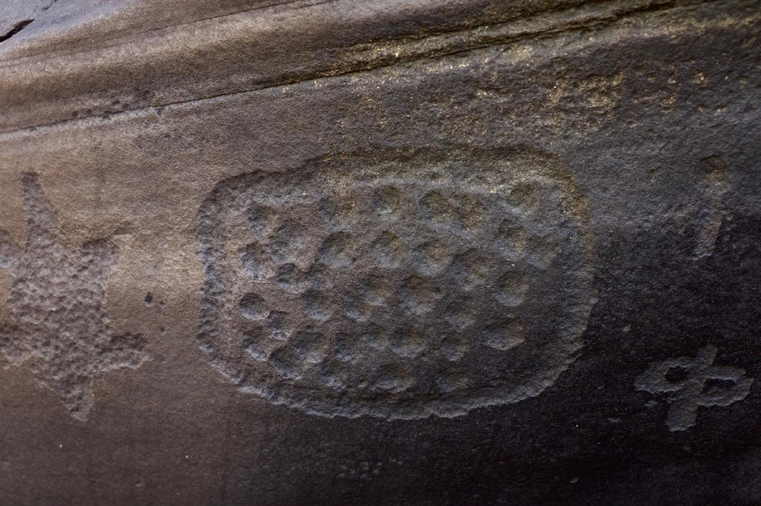}
\caption{}
\label{fig:sample22}
\end{subfigure}
\hfill
\begin{subfigure}{0.18\textwidth}
\centering
\includegraphics[width=\textwidth]{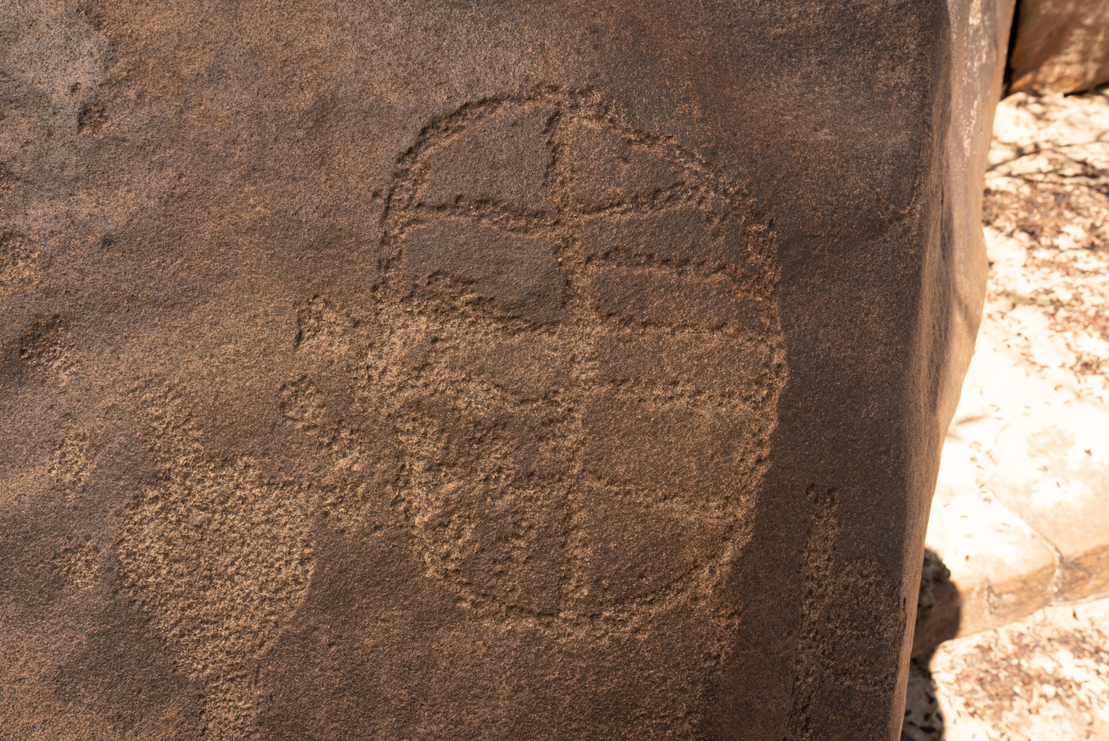}
\caption{}
\label{fig:sample45}
\end{subfigure}
\hfill
\begin{subfigure}{0.18\textwidth}
\centering
\includegraphics[width=\textwidth]{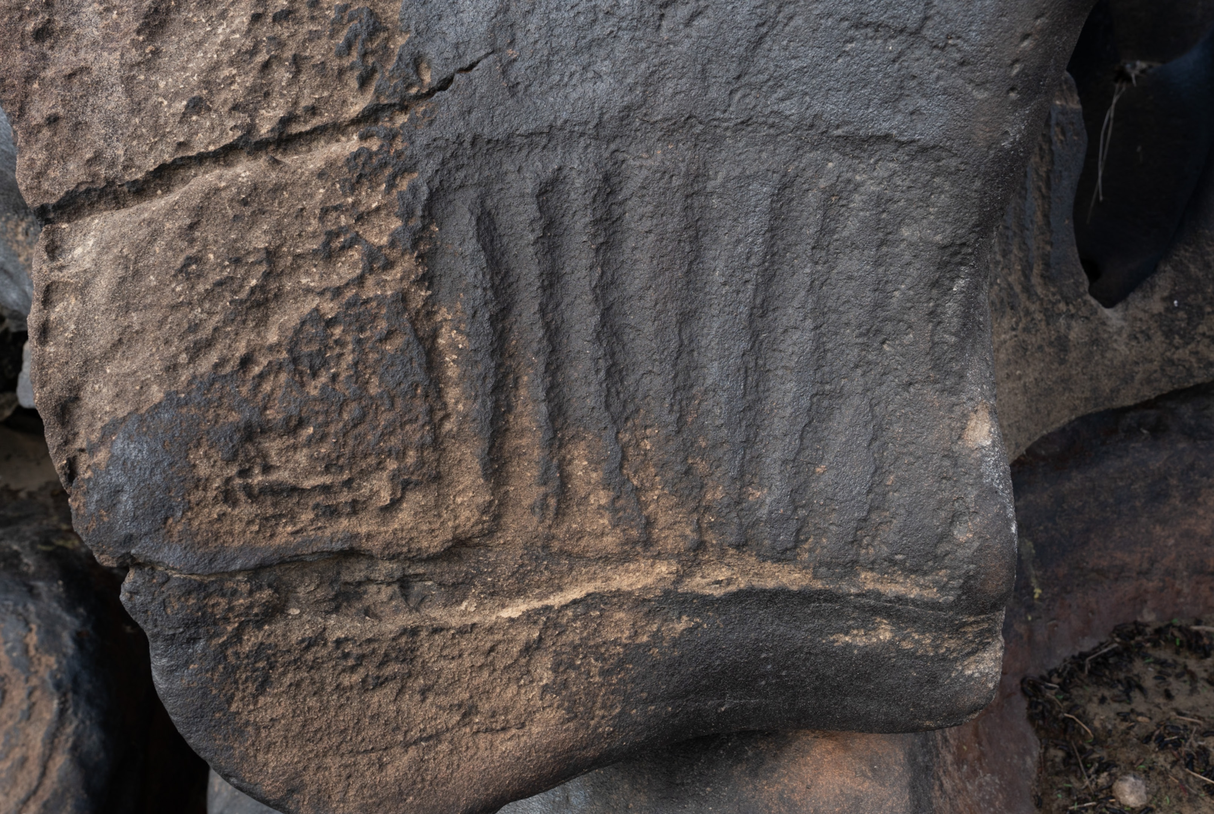}
\caption{}
\label{fig:sample67}
\end{subfigure}
\hfill
\begin{subfigure}{0.18\textwidth}
\centering
\includegraphics[width=\textwidth]{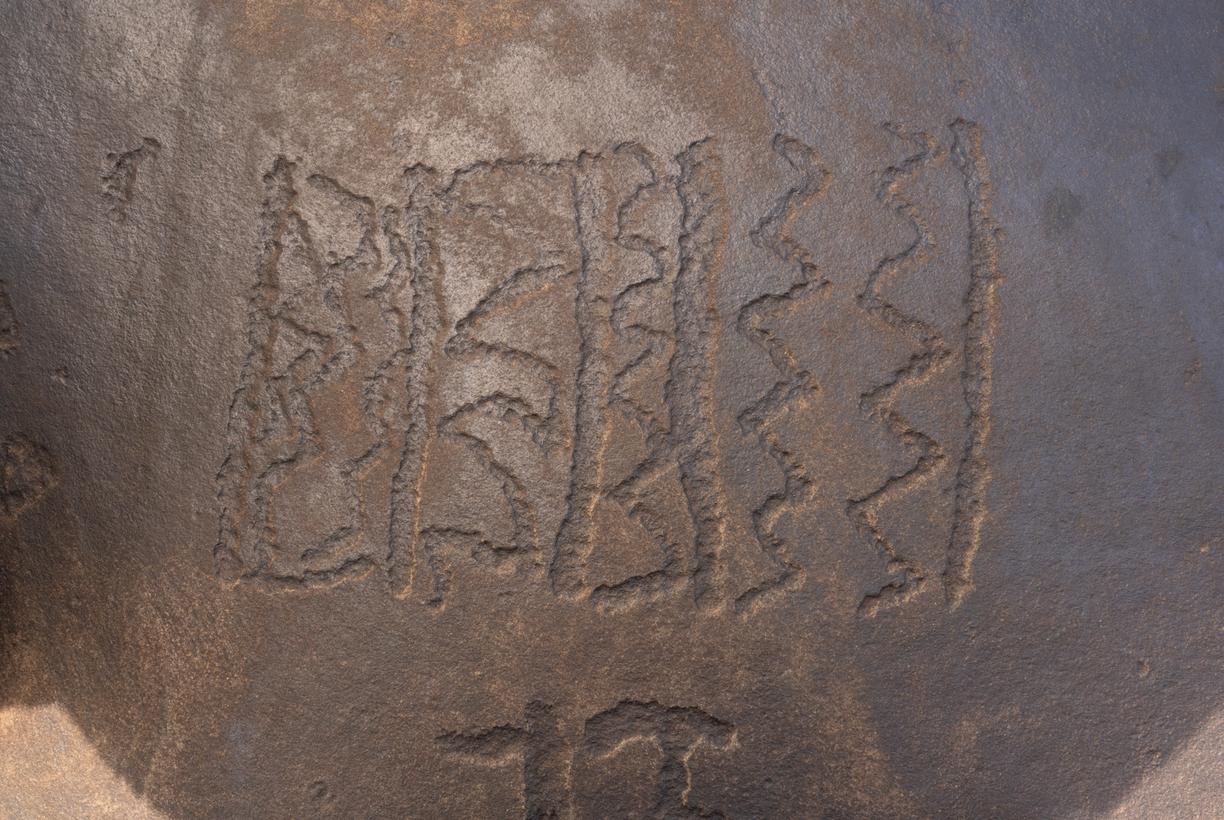}
\caption{}
\label{fig:sample78}
\end{subfigure}
\caption{Representative Dataset Samples of Brazilian Petroglyphs. Images show morphological variations including geometric patterns (\subref{fig:sample10}), linear engravings (\subref{fig:sample22}), complex overlapped figures (\subref{fig:sample45}), anthropomorphic representations (\subref{fig:sample67}), and zoomorphic figures with differential erosion (\subref{fig:sample78}).}
\label{fig:dataset_samples}
\end{figure}

Figure \ref{fig:dataset_samples} documents morphological variations present in the dataset, including different petroglyph typologies, conservation states, and geometric patterns.

\subsection{Implemented Architectures}

\subsubsection{BEGL-UNet (Baseline)}

The baseline implementation follows the original U-Net specification \cite{ronneberger2015}. The encoder path consists of four hierarchical levels with double convolutional blocks following channel progression 3→64→128→256→512. Each block contains two sequential 3×3 convolutions with batch normalization and ReLU activations. Max pooling operations with 2×2 kernels are applied between encoder levels to progressively reduce spatial resolution and capture multi-scale features. At the bottleneck, a convolutional block at minimum resolution (32×32 pixels) expands to 1024 channels, representing the highest level of feature abstraction where complex patterns are processed. The decoder path utilizes transposed convolutions with 2×2 kernels and stride 2 for progressive upsampling, restoring spatial resolution with sequential channel reduction: 1024→512→256→128→64. Direct concatenation between encoder and decoder features at corresponding levels preserves high-resolution spatial information along the channel dimension through skip connections. The output layer applies a final 1×1 convolution reducing to a single channel, followed by sigmoid activation: $output = \sigma(\text{Conv}_{64 \rightarrow 1}(x))$, mapping values to the [0,1] range appropriate for binary segmentation.

%%%%%% Attention-Residual
\subsubsection{Attention-Residual BEGL-UNet}
The Attention-Residual BEGL-UNet architecture integrates residual blocks with attention mechanisms and BEGL loss function for rock art segmentation. The implementation employs post-activation residual blocks based on He et al. \cite{he2016} with classical design (Conv-BN-ReLU-Conv-BN) followed by residual connection, formulated as:
\begin{align}
\mathbf{y} = \text{ReLU}(\mathcal{F}(\mathbf{x}, \{W_i\}) + \mathbf{x}) \label{eq:residual_postact}
\end{align}
where $\mathcal{F}$ represents the residual mapping implemented with identity connections for different channels via 1×1 convolution. Gated attention mechanisms are applied at each decoder level for suppression of irrelevant features, defined by:
\begin{align}
g_1 &= \text{Conv}_{1 \times 1}(\text{BN}(g)) \\
x_1 &= \text{Conv}_{1 \times 1}(\text{BN}(x)) \\
\psi &= \sigma(\text{Conv}_{1 \times 1}(\text{BN}(\text{ReLU}(g_1 + x_1)))) \\
\hat{x} &= x \odot \psi \label{eq:attention_gate_residual}
\end{align}
where $g$ represents the decoder gating signal, $x$ the encoder features, and $\odot$ denotes element-wise multiplication. The overall architecture follows a symmetric encoder-decoder structure with four hierarchical encoder levels (3→64→128→256→512 channels) and 1024-channel bottleneck, followed by decoder with upsampling via 2×2 transposed convolutions. The BEGL loss function combines binary cross-entropy with Gaussian edge enhancement:
\begin{align}
\mathcal{L}_{Residual-BEGL} = \alpha \cdot \text{MSE}(\mathcal{G}(\nabla pred), \mathcal{G}(\nabla target)) + \beta \cdot \text{BCE}(pred, target) \label{eq:residual_begl_loss}
\end{align}
with parameters $\alpha = 0.001$ and $\beta = 1.0$, where $\mathcal{G}$ represents Gaussian filter ($\sigma = 0.8$) and $\nabla$ Sobel operators for edge detection.

The architecture maintains skip connections from the original U-Net, integrating residual blocks in the encoding and decoding paths, with attention gates at concatenation points to focus on segmentation-relevant features.

%%%%% Spatial Channel Attention
\subsubsection{SCA-BEGL-UNet (Spatial-Channel Attention)}
This architecture integrates dual spatial-channel attention based on CBAM within hybrid gated attention blocks for rock art segmentation. The channel attention module employs CBAM implementation with 16:1 reduction ratio and parallel adaptive pooling processing, formulated as:
\begin{align}
\mathbf{M}_c &= \sigma(\mathbf{F}_{avg} + \mathbf{F}_{max}) \label{eq:channel_sca} \\
\mathbf{F}_{avg} &= \text{Conv}_{1 \times 1}(\text{ReLU}(\text{Conv}_{1 \times 1}(\text{AvgPool}(\mathbf{F})))) \nonumber \\
\mathbf{F}_{max} &= \text{Conv}_{1 \times 1}(\text{ReLU}(\text{Conv}_{1 \times 1}(\text{MaxPool}(\mathbf{F})))) \nonumber
\end{align}
where 1×1 convolutions replace fully connected layers to preserve spatial information. The spatial attention module processes channel statistics via dual pooling followed by 7×7 convolution:
\begin{align}
\mathbf{M}_s &= \sigma(\text{Conv}^{7\times7}([\text{AvgPool}_{channel}(\mathbf{F}); \text{MaxPool}_{channel}(\mathbf{F})])) \label{eq:spatial_sca}
\end{align}
where pooling operates along the channel dimension to create spatial attention maps. Hybrid attention gates integrate channel and spatial attention sequentially within gated attention blocks:
\begin{align}
g_1 &= \text{W}_g(g), \quad x_1 = \text{W}_x(x) \\
\psi_{base} &= \text{ReLU}(g_1 + x_1) \\
\psi_{channel} &= \text{ChannelAtt}(\psi_{base}) \odot \psi_{base} \\
\psi_{final} &= \text{SpatialAtt}(\psi_{channel}) \odot \psi_{channel} \label{eq:hybrid_attention_sca}
\end{align}
Residual blocks follow standard post-activation architecture with identity connections for different channel numbers via 1×1 convolution with batch normalization. The BEGL loss function follows standard implementation combining binary cross-entropy with Gaussian edge enhancement:
\begin{align}
\mathcal{L}_{SCA-BEGL} = \alpha \cdot \text{MSE}(\mathcal{G}(\nabla pred), \mathcal{G}(\nabla target)) + \beta \cdot \text{BCE}(pred, target) \label{eq:sca_begl_loss}
\end{align}
with parameters $\alpha = 0.001$ and $\beta = 1.0$, where $\mathcal{G}$ applies Gaussian filter ($\sigma = 0.8$) and $\nabla$ Sobel operators for edge detection.

The architecture integrates residual blocks and dual attention mechanisms, which operate complementarily by applying attention over spatial and channel features in attention gates to enhance edge localization in petroglyphs.

\subsection{BEGL Loss Function}

The BEGL loss was implemented following the specification of Bai et al. \cite{bai2023}, combining binary cross-entropy with Gaussian edge enhancement:

\begin{align}
\mathcal{L}_{BEGL} &= \alpha \mathcal{L}_G + \beta \mathcal{L}_{BCE} \label{eq:begl_main}
\end{align}

where the parameters used are $\alpha = 0.001$ and $\beta = 1.0$ for all architectures. The components are defined as:

\begin{align}
\mathcal{L}_{BCE} &= \text{BinaryCrossEntropy}(\hat{y}, y) \label{eq:bce_torch}\\
\mathcal{L}_G &= \text{MSE}(\mathcal{G}(\nabla \hat{y}), \mathcal{G}(\nabla y)) \label{eq:gaussian_loss}
\end{align}

The gradient operator $\nabla$ utilizes Sobel filters implemented via 2D convolution with padding=1:

\begin{align}
S_x &= \begin{bmatrix} -1 & 0 & 1 \\ -2 & 0 & 2 \\ -1 & 0 & 1 \end{bmatrix}, \quad
S_y = \begin{bmatrix} -1 & -2 & -1 \\ 0 & 0 & 0 \\ 1 & 2 & 1 \end{bmatrix} \label{eq:sobel}\\
|\nabla I| &= \sqrt{(S_x \ast I)^2 + (S_y \ast I)^2 + \epsilon} \label{eq:gradient}
\end{align}

where $\epsilon = 1 \times 10^{-8}$ prevents division by zero. The Gaussian filter $\mathcal{G}$ is implemented as a separable 2D kernel with parameters kernel\_size=5 and $\sigma = 0.8$:

\begin{align}
g[i] &= \frac{\exp(-(i - k//2)^2/(2\sigma^2))}{\sum_j \exp(-(j - k//2)^2/(2\sigma^2))} \\
\mathcal{G}(I) &= (g \otimes g^T) \ast I \label{eq:gaussian_separable}
\end{align}

where $k$ represents the kernel size and $\ast$ denotes 2D convolution with grouped channels for independent per-channel processing.

\subsection{Experimental Protocol}

\subsubsection{Data Augmentation}
\begin{figure}[h]
\centering
\begin{subfigure}{0.3\textwidth}
\centering
\includegraphics[width=\textwidth]{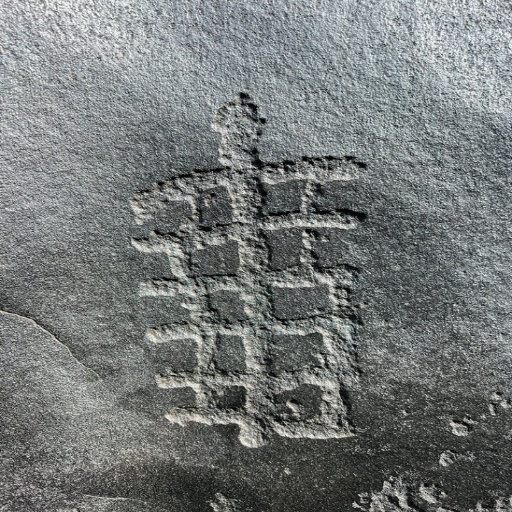}
\caption{}
\label{fig:aug_original}
\end{subfigure}
\hfill
\begin{subfigure}{0.3\textwidth}
\centering
\includegraphics[width=\textwidth]{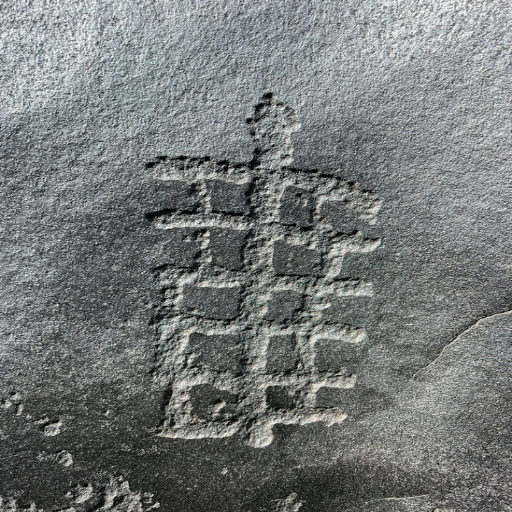}
\caption{}
\label{fig:aug_flip}
\end{subfigure}
\hfill
\begin{subfigure}{0.3\textwidth}
\centering
\includegraphics[width=\textwidth]{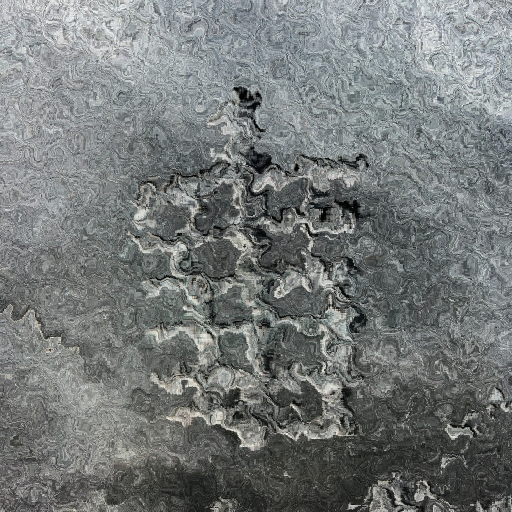}
\caption{}
\label{fig:aug_elastic}
\end{subfigure}
\caption{Data Augmentation Examples. Representative transformations applied to training set: (\subref{fig:aug_original}) original image, (\subref{fig:aug_flip}) horizontal flip transformation, (\subref{fig:aug_elastic}) elastic deformation with parameters alpha=120 and sigma=6.}
\label{fig:augmentation_examples}
\end{figure}

Data augmentation techniques were implemented via the Albumentations library \cite{albumentations2020}. Geometric transformations included horizontal and vertical flips (p=1.0), while elastic deformations were applied with parameters $\alpha=120$ and $\sigma=6$ (p=1.0). Optical distortions utilized distortion limit=2 and shift limit=0.5 (p=1.0), combined with grid distortion (p=1.0). A 6× augmentation factor was applied to the training set, whereas test set samples were maintained without modifications for validation. Figure \ref{fig:augmentation_examples} presents applied transformations, demonstrating preservation of morphological characteristics of petroglyphs.

\subsubsection{Training Configuration}
The learning rate was set to $\eta = 1 \times 10^{-4}$. Adam optimizer was employed with parameters $\beta_1 = 0.9$ and $\beta_2 = 0.999$. Due to memory constraints, batch size was limited to $B = 2$. BEGL loss parameters were configured as $\alpha = 0.001$ and $\beta = 1.0$, with weight decay $\lambda = 1 \times 10^{-6}$.

\subsubsection{Implementation Details}
All models were implemented in PyTorch and trained on an NVIDIA RTX 4070 Super GPU with 12GB VRAM. Random seeds were fixed at 42 for NumPy, PyTorch, and CUDA to ensure reproducibility. Training duration was approximately 3-4 hours per architecture for 50 epochs with the specified batch size. The implementation utilized standard PyTorch libraries including torch.nn for network layers, torch.optim for optimization, and scikit-learn for cross-validation and metric computation.

\section{RESULTS}

\subsection{Quantitative Analysis}

Table \ref{tab:results_final} presents consolidated results from 5-fold cross-validation for all three architectures.

\begin{table}[!htb]
\caption{Architecture Performance Comparison (5-fold Cross-Validation Results)}
\label{tab:results_final}
\centering
\begin{tabular}{lcccccc}
\hline
\textbf{Architecture} & \textbf{Val Loss} & \textbf{DSC} & \textbf{Precision} & \textbf{Recall} & \textbf{F1-score} & \textbf{Pixel Acc} \\
\hline
BEGL-UNet & $0.077$ & $0.690$ & $0.617$ & $0.834$ & $0.690$ & $0.968$ \\
Attention-Residual BEGL-UNet & $\mathbf{0.067}$ & $\mathbf{0.710}$ & $\mathbf{0.629}$ & $\mathbf{0.854}$ & $\mathbf{0.710}$ & $\mathbf{0.967}$ \\
Spatial-Channel Attention BEGL-UNet & $0.081$ & $0.707$ & $0.618$ & $0.857$ & $0.707$ & $0.968$ \\
\hline
\multicolumn{7}{l}{\footnotesize All values represent mean performance across 5 folds at epoch 30.} \\
\end{tabular}
\end{table}

\subsubsection{Comparative Performance}

Attention-Residual BEGL-UNet achieved the best overall performance with a DSC of 0.710 compared to the baseline DSC of 0.690, representing a 2.9\% improvement. This architecture also obtained the lowest validation loss (0.067) and highest precision (0.629), recall (0.854), and F1-score (0.710). Spatial-Channel Attention BEGL-UNet achieved DSC of 0.707, representing a 2.5\% improvement over baseline, with comparable recall (0.857) to the Attention-Residual architecture. All architectures achieved pixel accuracy of approximately 96.7-96.8%. The Attention-Residual BEGL-UNet demonstrated the most consistent performance across all evaluated metrics.

\subsection{Qualitative Analysis}

\begin{figure}[h]
\centering
\begin{subfigure}{0.3\textwidth}
\centering
\includegraphics[width=\textwidth]{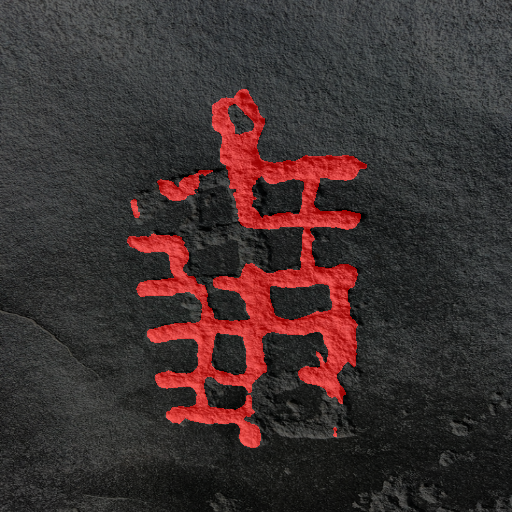}
\caption{}
\label{fig:begl_baseline}
\end{subfigure}
\hfill
\begin{subfigure}{0.3\textwidth}
\centering
\includegraphics[width=\textwidth]{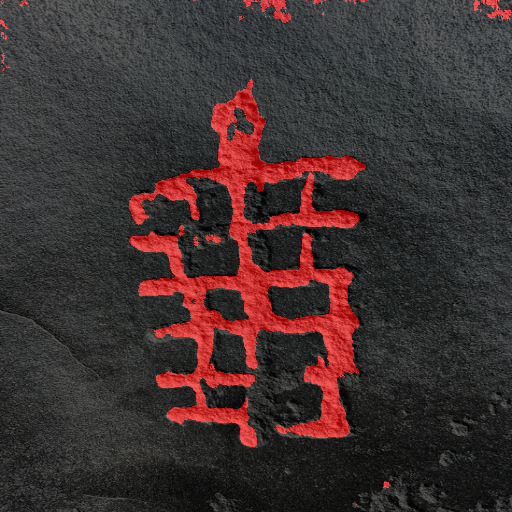}
\caption{}
\label{fig:attention_residual}
\end{subfigure}
\hfill
\begin{subfigure}{0.3\textwidth}
\centering
\includegraphics[width=\textwidth]{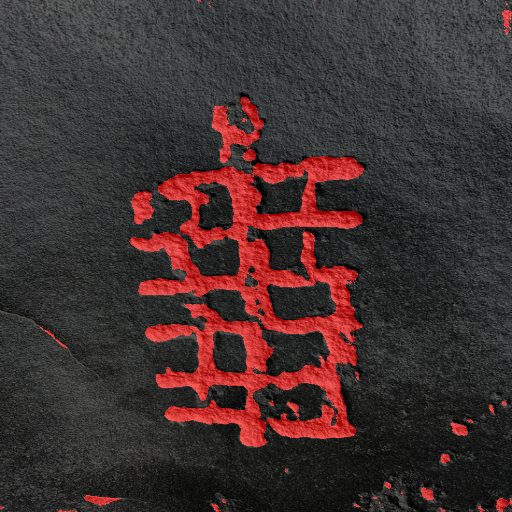}
\caption{}
\label{fig:sca_begl}
\end{subfigure}
\caption{Comparative Segmentation Results on Sample 15. Progressive improvement in edge definition and morphological accuracy across architectures: (\subref{fig:begl_baseline}) BEGL-UNet baseline, (\subref{fig:attention_residual}) Attention-Residual BEGL-UNet demonstrating superior boundary delineation, (\subref{fig:sca_begl}) Spatial Channel Attention BEGL-UNet with comparable edge precision.}
\label{fig:comparative_segmentation}
\end{figure}

Figure \ref{fig:comparative_segmentation} presents comparative segmentation results of all three architectures on petroglyph sample 15, demonstrating visual differences in edge precision and morphological preservation.

\subsection{Convergence Analysis}

All models demonstrated convergence by epoch 30 of training. The Attention-Residual BEGL-UNet achieved the lowest validation loss (0.067) with stable performance across folds. Spatial-Channel Attention BEGL-UNet and BEGL-UNet baseline presented validation losses of 0.081 and 0.077 respectively. The consistency across 5-fold cross-validation indicates that the architectures achieved stable training despite the limited dataset size, with attention mechanisms providing consistent performance improvements.

\subsection{Failure Cases Analysis}
\begin{figure}[!htbp]
\centering
\begin{subfigure}{0.3\textwidth}
\centering
\includegraphics[width=\textwidth]{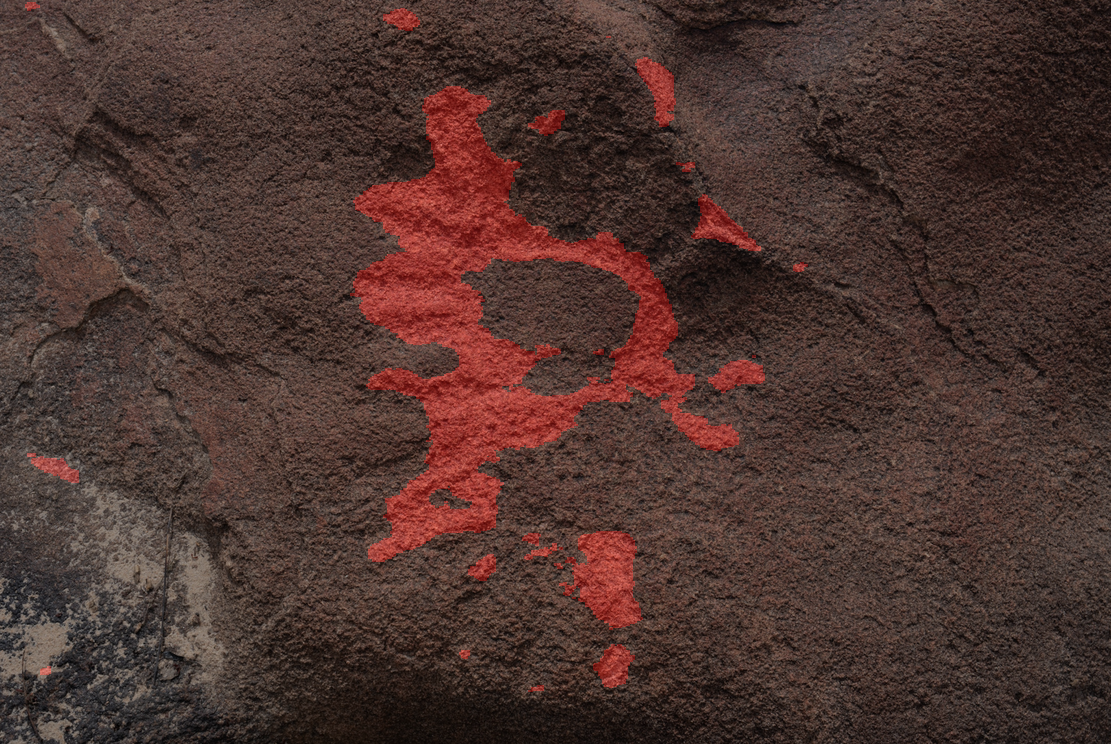}
\caption{}
\label{fig:failure_superposition}
\end{subfigure}
\hfill
\begin{subfigure}{0.3\textwidth}
\centering
\includegraphics[width=\textwidth]{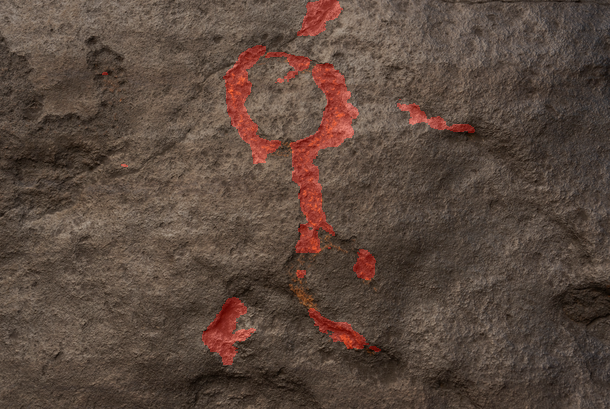}
\caption{}
\label{fig:failure_minerals}
\end{subfigure}
\hfill
\begin{subfigure}{0.3\textwidth}
\centering
\includegraphics[width=\textwidth]{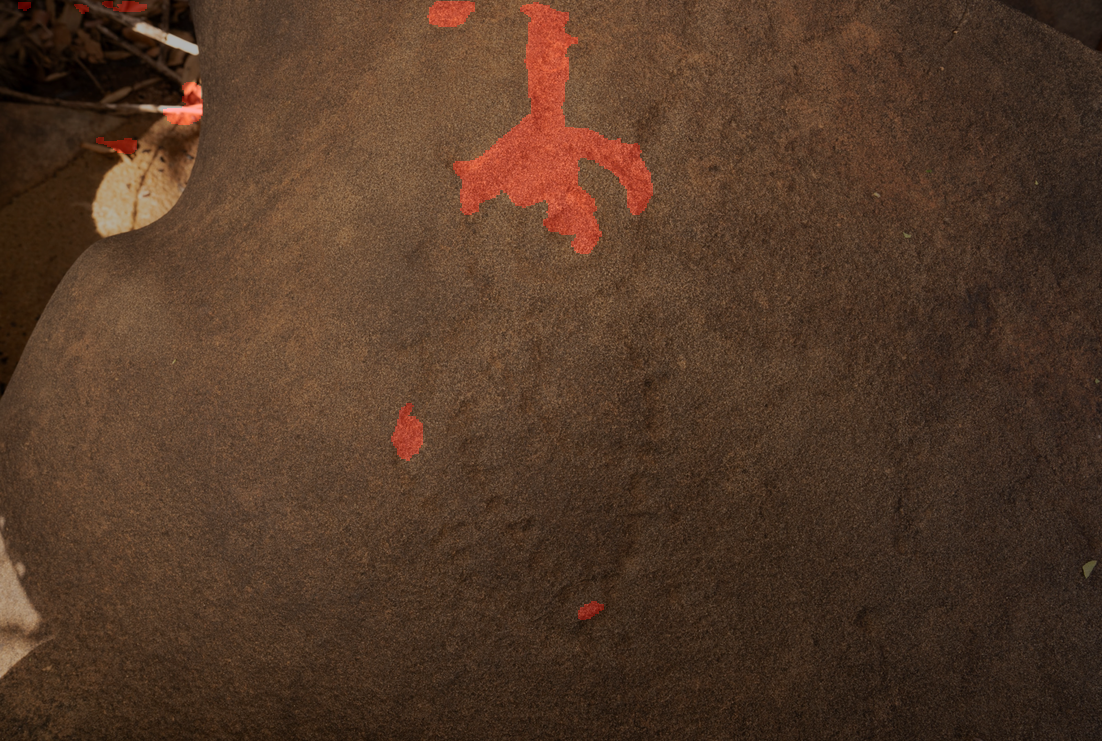}
\caption{}
\label{fig:failure_contrast}
\end{subfigure}
\caption{Representative Failure Cases. Challenging conditions affecting segmentation performance across all three architectures: (\subref{fig:failure_superposition}) chronological superposition with multiple temporal phases of engraving creating ambiguous boundaries, (\subref{fig:failure_minerals}) mineral deposits and iron oxide crusts occluding petroglyphs, (\subref{fig:failure_contrast}) low radiometric contrast between engraving and gneiss substrate under uncontrolled illumination.}
\label{fig:failure_cases}
\end{figure}

Results analysis identified specific conditions where all architectures exhibited inferior performance. These limitations affected all three evaluated architectures equally across challenging samples. Figure \ref{fig:failure_cases} illustrates the main failure categories observed. Main observed limitations include:

\textbf{Chronological superposition}: Engravings executed in multiple temporal phases created superpositions with ambiguous boundaries. Spatial-Channel Attention BEGL-UNet achieved the highest recall (0.857), indicating improved detection capability in complex scenarios.

\textbf{Mineral deposition and biological colonization}: Formation of iron oxide crusts and colonization by lichen organisms partially occluded rock art patterns. All architectures presented similar challenges in these cases, with modest overall DSC values reflecting the difficulty of this segmentation task.

\textbf{Low radiometric contrast}: Variability in mineralogical composition of the gneiss substrate and uncontrolled illumination conditions resulted in limited spectral differentiation. Attention-Residual BEGL-UNet obtained the best precision (0.629) and balanced recall (0.854), demonstrating more robust performance across challenging conditions.

\section{DISCUSSION}

\subsection{Performance Analysis}

The three analyzed architectures exhibited performance variation in the segmentation task. BEGL-UNet baseline registered DSC of 0.690 and validation loss of 0.077. Attention-Residual BEGL-UNet obtained the best overall results with DSC of 0.710 and validation loss of 0.067, achieving the highest precision (0.629) and recall (0.854) among all architectures. Spatial-Channel Attention BEGL-UNet presented comparable DSC of 0.707 with slightly higher recall (0.857) but validation loss of 0.081. All architectures achieved pixel accuracy of approximately 96.7-96.8\%. These results indicate that attention mechanisms provide modest but consistent improvements over the baseline, with the Attention-Residual architecture demonstrating the most balanced performance across metrics.

\subsection{Practical Implications for Archaeological Applications}

The Attention-Residual BEGL-UNet model (DSC = 0.710, pixel accuracy = 96.7\%) presents characteristics relevant to archaeological documentation workflows. The balanced precision (0.629) and recall (0.854) with low validation loss (0.067) indicate consistent performance across different petroglyph morphologies. While the absolute DSC values are moderate, the improvements over baseline demonstrate the potential of attention mechanisms for rock art segmentation in challenging field conditions with variable illumination and weathering patterns.

\section{CONCLUSION}

This study evaluated U-Net architectures with attention mechanisms for semantic segmentation of rock art on a Brazilian petroglyph dataset. Attention-Residual BEGL-UNet presented the best overall performance, with DSC of 0.710, validation loss of 0.067, and recall of 0.854, representing a 2.9\% improvement over the baseline. Spatial-Channel Attention BEGL-UNet achieved comparable performance with DSC of 0.707 and recall of 0.857. Both attention-enhanced architectures surpassed the baseline BEGL-UNet (DSC: 0.690), demonstrating the effectiveness of attention mechanisms for rock art segmentation.

\subsection{Limitations and Future Work}
The primary limitation of this study relates to the utilization of a dataset restricted to a single archaeological site, which may limit generalization of results to other contexts. Research perspectives include evaluation of Vision Transformers for spatial dependency modeling, development of hierarchical multi-scale processing strategies, exploration of self-supervised learning, expansion of the dataset to different Brazilian archaeological sites, automatic estimation of petroglyph dimensions, directional analysis of engraving techniques, and integration with 3D reconstruction methods for archaeological documentation.

\section*{Acknowledgments}

\subsection*{General}
The authors acknowledge Dr. Gildário Lima and Prof. Lucciani Vieira for the guidance provided and for proposing this study. Institutional support from iCEV Institute of Higher Education is also recognized.

\subsection*{Author Contributions}
L. Melo and L. Gustavo were responsible for the conception, implementation, and execution of the entire study. D. Magalhães, L. Vieira, and M. Araújo supervised the research and provided guidance throughout the project. All authors contributed equally to the preparation and revision of the manuscript.

%%%%%% Insert references here %%%%%%%%%%%%%%%%%%%%%
\bibliographystyle{plain}
\bibliography{references}

\end{document}